\documentclass[sigconf]{acmart}

\AtBeginDocument{%
  }

\copyrightyear{2024}
\acmYear{2024}
\setcopyright{acmlicensed}\acmConference[CIKM '24]{Proceedings of the 33rd ACM International Conference on Information and Knowledge Management}{October 21--25, 2024}{Boise, ID, USA}
\acmBooktitle{Proceedings of the 33rd ACM International Conference on Information and Knowledge Management (CIKM '24), October 21--25, 2024, Boise, ID, USA}
\acmDOI{10.1145/3627673.3679541}
\acmISBN{979-8-4007-0436-9/24/10}

\usepackage{xspace}
\usepackage{subcaption}
\usepackage{float}
\usepackage{algorithm}
\usepackage{algorithmic}

\usepackage{titlesec}
\usepackage{balance}

\newcommand\methodname{HYPER-MGE\xspace}
\newcommand{\norm}[1]{\left\lVert#1\right\rVert}

\begin{document}

\title{A Geometric Perspective for High-Dimensional Multiplex Graphs}


\author{Kamel Abdous}
\affiliation{%
  \department{Department of Computer Science}
  \institution{University of Quebec at Montreal}
  \city{Montreal}
  \state{Quebec}
  \country{Canada}
}
\email{abdous.kamel@courrier.uqam.ca}

\author{Nairouz Mrabah}
\affiliation{%
  \department{Department of Computer Science}
  \institution{University of Quebec at Montreal}
  \city{Montreal}
  \state{Quebec}
  \country{Canada}
}
\email{mrabah.nairouz@courrier.uqam.ca}

\author{Mohamed Bouguessa}
\affiliation{%
  \department{Department of Computer Science}
  \institution{University of Quebec at Montreal}
  \city{Montreal}
  \state{Quebec}
  \country{Canada}
}
\email{bouguessa.mohamed@uqam.ca}


\begin{abstract}
High-dimensional multiplex graphs are characterized by their high number of complementary and divergent dimensions. The existence of multiple hierarchical latent relations between the graph dimensions poses significant challenges to embedding methods. In particular, the geometric distortions that might occur in the representational space have been overlooked in the literature. This work studies the problem of high-dimensional multiplex graph embedding from a geometric perspective. We find that the node representations reside on highly curved manifolds, thus rendering their exploitation more challenging for downstream tasks. Moreover, our study reveals that increasing the number of graph dimensions can cause further distortions to the highly curved manifolds. To address this problem, we propose a novel multiplex graph embedding method that harnesses hierarchical dimension embedding and Hyperbolic Graph Neural Networks. The proposed approach hierarchically extracts hyperbolic node representations that reside on Riemannian manifolds while gradually learning fewer and more expressive latent dimensions of the multiplex graph. Experimental results on real-world high-dimensional multiplex graphs show that the synergy between hierarchical and hyperbolic embeddings incurs much fewer geometric distortions and brings notable improvements over state-of-the-art approaches on downstream tasks.
\end{abstract}



\maketitle

\section{Introduction}

Recent years have witnessed the advent of complex real-world systems, where basic units are connected via multiple types of interaction. In this context, multiplex graphs have gained popularity for representing the interdependent structure of these systems \cite{kivela2014multilayer}. Multiplex graphs are characterized by their multidimensionality; the connectivity of the nodes differs from one dimension to another. Among the principal concepts that govern this type of data is the existence of complementary and divergent information across different dimensions \cite{de2015structural}. This applies in particular when the number of dimensions increases, leading to the emergence of high-dimensional multiplex graphs. The divergence across a multitude of dimensions presents significant challenges for embedding algorithms. More precisely, high-dimensional multiplex graph embedding methods should uncover convoluted relations across diverse dimensions to extract a unified graph structure and more informative node representations.

Most multiplex graph embedding techniques rely on two strategies: Random Walks (RWs) and Graph Neural Networks (GNNs). RW-based methods \cite{cen2019representation,pio2021multiverse} generate sequences of nodes on individual dimensions to extract dimension-specific embeddings, which are then linearly aggregated into consensus embeddings. In the same way, several GNN-based approaches \cite{park2020dmgi,jing2021hdmi,mitra2021semi} produce dimension-specific node embeddings and then perform a linear aggregation of these embeddings. However, both strategies fail to consider the hierarchical relations between the graph dimensions, which cannot be captured by a single linear aggregation. Recently, a first attempt has been made to address this critical issue. The authors of \cite{hmge2023} have introduced a hierarchical aggregation strategy that summarizes the graph dimensions into gradually smaller sets. However, the impact of existing multiplex graph embedding techniques, including the hierarchical aggregation strategy, on the embedding space geometry remains unexplored. In particular, it is important to investigate the effect of increasing the number of dimensions from a geometric perspective.

An interesting geometric perspective for understanding multiplex graph embedding is to assess the Intrinsic Dimension (ID) and the Linear Intrinsic Dimension (LID) of the latent manifolds. On the one hand, the ID measures the minimal number of variables needed to describe the data. On the other hand, the LID measures the dimension of the smallest subspace that can enclose all data points. The difference between ID and LID highlights how much the data structure deviates from a linear subspace. When the LID exceeds the ID, it implies that the manifold has a curved structure, which cannot be effectively represented by a linear subspace with the same dimension \cite{ansuini2019intrinsic}. The greater the difference, the more curved the manifold will be. Using the ID and LID estimations, it has been shown that training vanilla neural networks in a supervised fashion transforms the initial latent structures into low-dimensional and curved ones \cite{ansuini2019intrinsic}. In another work, the authors of \cite{mrabahescaping} have shown that training graph neural networks for uni-dimensional graphs under the deep clustering paradigm causes abrupt geometric distortions of the latent manifolds after the pretraining phase. However, no previous work has studied the evolution of ID and LID in the context of multiplex graph embedding. In this work, we focus on answering two questions: What are the geometric implications of embedding diverse structural patterns spread across multiple graph dimensions? How to encode the hierarchical relations between the dimensions of a multiplex graph without causing significant geometric distortions?

To answer the first question, we conduct a geometric investigation. First, we construct synthetic multiplex graphs that span a large spectrum of dimensions, with node clusters spread across multiple dimensions. Using synthetic data allows to control the number of dimensions and the extent of divergent information (i.e., between-cluster connections) across the graph structures. Then, we assess the geometric distortions by measuring the discrepancy between the ID and LID for state-of-the-art embedding strategies. Our results suggest that the encoding process subjects the latent space to coarse transformations. More precisely, we find that the node embeddings reside on highly-curved and low-dimensional manifolds independently of the number of graph dimensions. Moreover, we observe that increasing the number of graph dimensions, which is accompanied by an increase in divergent information (i.e., presence of between-cluster connections across all dimensions) and a dilution of relevant information (i.e., sparser within-cluster connections spread across several dimensions), strengthens the geometric distortions and makes the latent manifolds more curved. Consequently, the learned representations under these geometric distortions are suboptimal for downstream tasks.

Recently, Hyperbolic Graph Neural Networks (HGNNs) \cite{chami2019hyperbolic,liu2019hyperbolic} have been devised to encode uni-dimensional graphs with complex topologies (e.g., hierarchical and looping structures \cite{nickel2017poincare}). HGNNs project graph nodes into Riemannian manifolds, such as Poincaré balls and spheres \cite{liu2019hyperbolic}. This process generates hyperbolic representations, which are well suited for solving downstream tasks on graphs with tree-like structural patterns \cite{bachmann2020constant,sun2023contrastive}. To answer the second question raised in this work, 
we elaborate a hyperbolic-based embedding approach to tackle the geometric distortions inherent in multiplex graph encoding methods. We find that hyperbolic embedding, coupled with hierarchical dimension aggregation, not only captures improved representations of high-dimensional graphs but also yields latent manifolds with minimal geometric distortions.

\subsection*{Contributions}
\textbf{(i) Identification of a new problem:} One of the core contributions of our work lies in identifying the geometric distortions that occur in the embedding of high-dimensional multiplex graphs. More precisely, we show that increasing the number of graph dimensions, which is accompanied by an increase in divergent information (i.e., more between-cluster connections across all dimensions) and a dilution of relevant information (i.e., sparser within-cluster connections spread across several dimensions), leads to the occurrence of geometric distortions and makes the latent manifolds more curved. We provide a geometric study of these distortions on synthetic and real-world graphs. \textbf{(ii) Methodological novelty:} We propose \methodname, a novel approach that can learn hierarchical representations of the graph dimensions by embedding them into hyperbolic spaces. We argue that hierarchical aggregations can capture complex hidden structures and that hyperbolic geometry can account for the hierarchical latent structures without geometric distortions. \textbf{(iii) Empirical validation:}  We support our claims with extensive experiments that show the effectiveness of our approach in reducing geometric distortions and improving performance on downstream tasks compared to existing methods. Our approach brings substantial enhancement over the state-of-the-art methods in several cases.

\section{Related Work}

Consistent with our focus, we examine multiplex graph embedding and hyperbolic graph neural network techniques.

\subsection{Multiplex Graph Embedding}

MultiVERSE \cite{pio2021multiverse} learns node representations from random walks that traverse the graph. To handle the multidimensional aspect of multiplex graphs, the constructed node sequences can transit from one dimension to another. Similarly, GATNE \cite{cen2019representation} converts random walks into training samples and uses trainable transformations to compute node embeddings. In a high-dimensional setting, random walk-based methods require long sequences of nodes to cover all the dimensions of a multiplex graph. Besides, local optimization algorithms (such as Skip-Gram \cite{mikolov2013distributed}) form the basis of these methods' training modules. Thus, long-range dependencies spanning multiple dimensions can be difficult to identify. Last but not least, random walk-based methods lack the expressive power to capture compositional and convoluted relations across diverse dimensions of high-dimensional multiplex graphs.

Another line of work exploits Graph Neural Networks (GNNs) to encode multiplex graphs into low-dimensional vectors. DMGI \cite{park2020dmgi} and HDMI \cite{jing2021hdmi} apply multiple GNNs to learn node embeddings on individual dimensions. Then, a linear aggregation step with an attention mechanism converts the dimension-specific embeddings into consensus codes. Both models, DMGI and HDMI, optimize a contrastive loss based on mutual information maximization \cite{velickovic2019deep}, but HDMI has a higher-order objective function that includes node features. X-GOAL \cite{jing2022x} improves the contrastive loss by grouping topologically similar nodes and nodes within the same cluster into positive and negative pairs. SSDCM \cite{mitra2021semi} maximizes the mutual information between local node-level representations and a global cluster-aware graph summary. DMG \cite{mo2023disentangled} and MGDCR \cite{mgdcr} focus on the common and complementary information among the multiplex graph dimensions. DMG employs disentangled representations to separate between common and private information. MGDCR uses a loss function to minimize the correlation between inter-dimension and intra-dimension representations. In contrast with the unsupervised methods considered in this paper, SSAMN \cite{ssamn2023} is a semi-supervised approach that integrates spectral embedding to encode multiplex graphs with a small number of labeled nodes provided during training. Overall, all these methods fail to consider the \textit{compositional} nature of high-dimensional multiplex graphs and are thus hindered by information loss caused by the single and linear aggregation step. 

The authors of HMGE \cite{hmge2023} have established the existence of hierarchical relations between the dimensions of real-world multiplex graphs. This concept describes how new high-level graph dimensions can be formed from the non-linear hierarchical combinations of lower-level initial dimensions. Motivated by this finding, HMGE introduces a hierarchical aggregation mechanism instead of the single and linear aggregation. More precisely, HMGE defines trainable non-linear combinations stacked in a gradual way to shrink the number of dimensions and build new relevant ones. However, previous GNN-based methods, including HMGE, overlook the geometric distortions that might occur in the latent space due to the hierarchical relations between a high number of graph dimensions. In this work, we establish the presence of these geometric distortions at the latent space and propose a solution to this problem based on hyperbolic embedding.

\subsection{Hyperbolic Graph Neural Networks}

These models map the input graph into hyperbolic spaces. The key aspect of a hyperbolic space is its exponential volume expansion with respect to its radius, whereas an Euclidean space exhibits a polynomial growth in volume. As a result, HGNNs can effectively encode complex patterns, including hierarchical and looping structures \cite{chami2019hyperbolic,zhu2020graph}. This is because a hyperbolic space aligns with the growth rate of tree-like patterns, a property that an Euclidean space can not grant. HGNNs produce high-quality representations in real-world scenarios and can improve performance compared to Euclidean competitors in various downstream tasks \cite{peng2021hyperbolic}.

From a geometric perspective, previous works \cite{xiong2023geometric} have shown that embedding unidimensional graphs with complex structural patterns causes geometric distortions in the latent space. To alleviate this problem, hyperbolic embedding techniques are known to reduce these distortions for graphs with tree-like structural patterns \cite{bachmann2020constant}. In practice, the most prevalent hyperbolic manifolds correspond to the Poincaré and Lorentz models \cite{liu2019hyperbolic}. Although HGNNs have shown success in capturing hierarchical relations between the graph nodes with minimal geometric distortions, it is still unknown if embedding the hierarchical relations between the graph dimensions can cause geometric distortions in the latent space. If so, it is important to provide a solution for encoding these hierarchical relations in a way that exhibits minimal geometric distortions.

\section{A Geometric Study of High-Dimensional Multiplex Graph Embedding}\label{seq:geo_study}

In this section, we identify a new problem that affects multiplex graph embedding methods. Specifically, we answer the question: \textit{What are the geometric implications of embedding multiple graph dimensions?} To this end, we conduct an empirical study that investigates the geometric transformations on the latent manifolds caused by embedding high-dimensional multiplex graphs. We devise a two-phase protocol: \textbf{(i)} Generating synthetic multiplex graphs that span a large spectrum of dimensions and simulate the structure of real-world data. \textbf{(ii)} Training state-of-the-art multiplex graph embedding methods on the synthetic data and monitoring the evolution of the ID and LID metrics as the number of dimensions increases to assess the distortions.

\subsection{Synthetic Data Generation}

Real-world multiplex graphs have two critical properties: \textbf{(i)} the abundance of between-cluster connections as the number of dimensions increases, and \textbf{(ii)} the spread of sparse within-cluster connections across several dimensions \cite{boutemine2017mining}. Accordingly, we devise a synthetic data generation process that simulates these two properties. Using synthetic data, we can \textit{control} the number of dimensions and the extent of divergent information (i.e., between-cluster connections) across the graph topological structures. This allows us to assess the impact of high dimensionality on the latent space geometry. Moreover, discarding the divergent information requires capturing complex hierarchical relations between the dimensions. This information would be hard to control without synthetic data. For these reasons, we generate synthetic multiplex graphs that span a high number of dimensions between $5$ and $100$. The full description of the multiplex graphs generation process is provided in Appendix \ref{app:synthetic_data}.

\subsection{Evaluation Protocol}\label{sec:geometric_study_protocol}

Our study includes state-of-the-art embedding approaches based on RWs and GNNs, such as DMGI \cite{park2020dmgi}, HDMI \cite{jing2021hdmi}, SSDCM \cite{mitra2021semi}, MultiVERSE \cite{pio2021multiverse}, X-GOAL \cite{jing2022x}, GATNE \cite{cen2019representation}, MGDCR \cite{mgdcr}, DMG \cite{mo2023disentangled}, and HMGE \cite{hmge2023}. We train the baselines on the synthetic datasets and measure the average ID and LID of the latent manifolds at the end of training. After that, we calculate the difference between the two metrics and plot the variation of this quantity with respect to the number of graph dimensions $D$. For a fair comparison, we set the size of the node embeddings to $64$ for all methods. The ID and LID metrics are described in Appendix \ref{app:intrinsic_dim}.

\subsection{Geometric Distortions in Latent Spaces}

\begin{figure}
\centering
\begin{subfigure}{1.0\linewidth}
    \centering
    \includegraphics[width=0.75\linewidth]{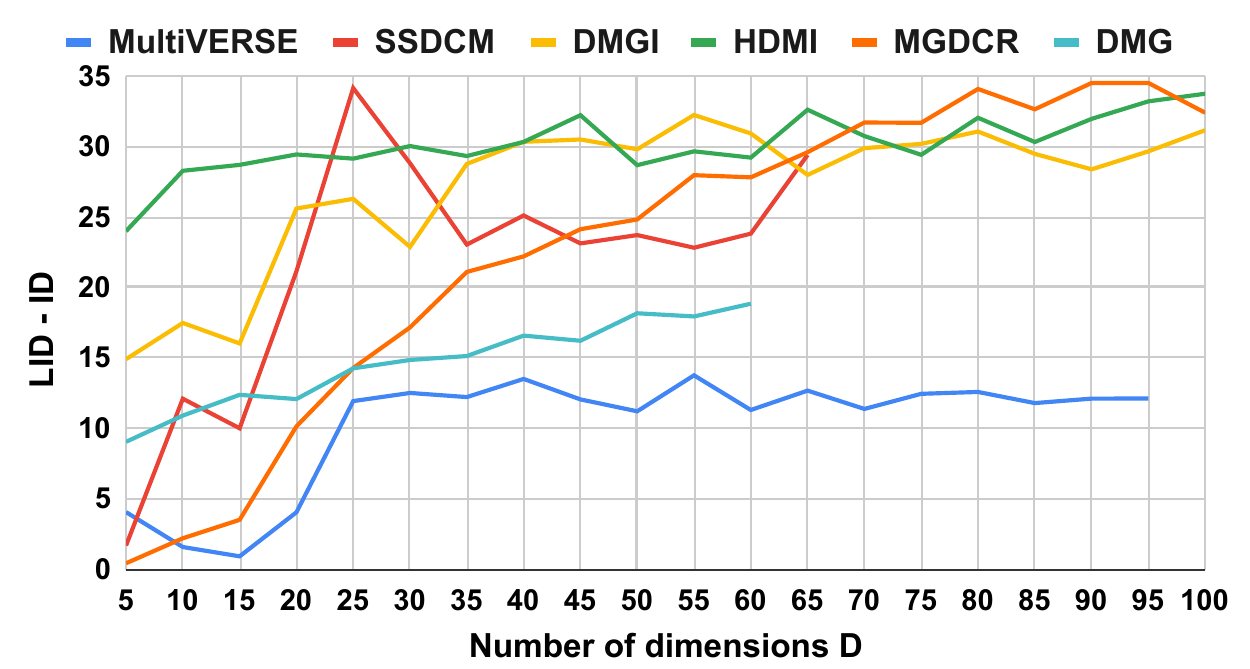}
    \caption{Increasing $D$ amplifies the disparity between the LID and ID.}
    \label{fig:cat1}
\end{subfigure}\\
\begin{subfigure}{1.0\linewidth}
    \centering
    \includegraphics[width=0.75\linewidth]{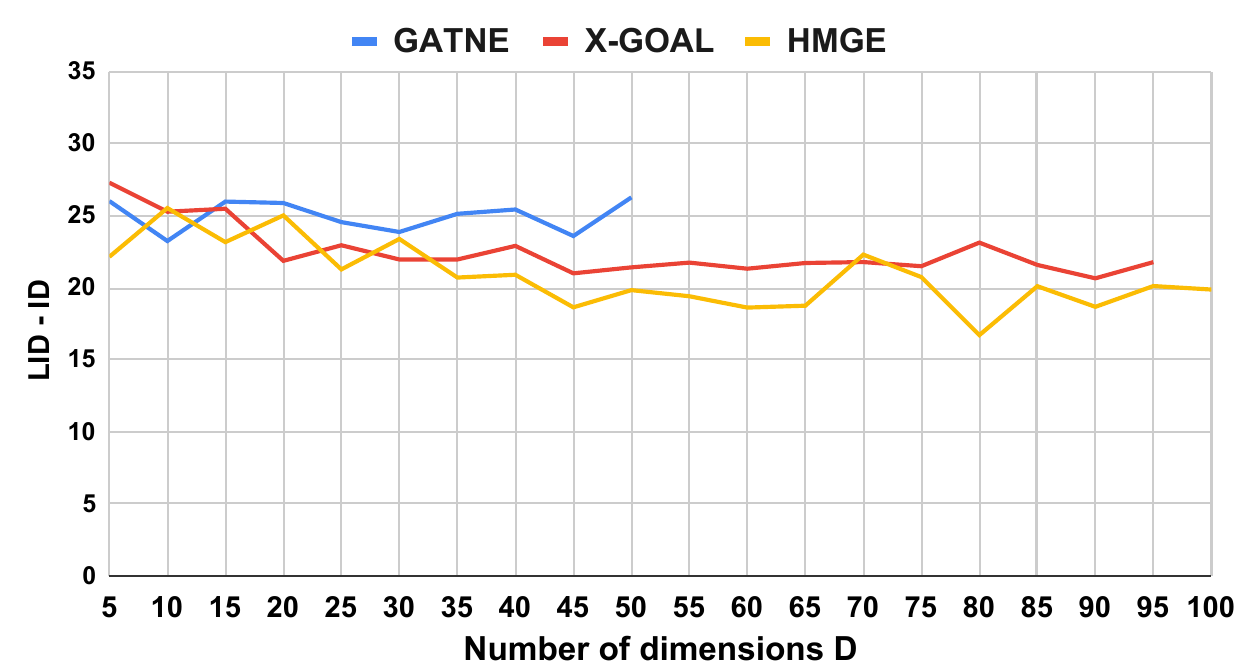}
   \caption{High difference between the LID and ID even when $D$ is small.}
    \label{fig:cat2}
\end{subfigure}

\caption{Results on synthetic high-dimensional multiplex graphs. Baselines such as SSDCM and GATNE lack measurements beyond some dimensions because they run out of memory.
}
\label{fig:geometrical_study}
\end{figure}

Figure \ref{fig:geometrical_study} shows the results of our geometric study. The horizontal axis corresponds to the number of dimensions in the synthetic multiplex graphs, while the vertical axis measures the difference between the average LID and ID of the latent manifolds at the end of the training process. It is worth noting that the difference between these metrics starts small for all baseline methods and consistently increases during training to peak by the end, regardless of the graph's dimension $D$. A \textit{large} difference between these two metrics indicates that the latent manifolds reach a \textit{highly-curved} state. We classify the approaches into two categories according to the results:

\textit{(1) The difference between the LID and ID increases as the number of dimensions $D$ increases:} We observe this behavior in multiple approaches (DMGI, HDMI, SSDCM, MGDCR, DMG, and MultiVERSE). For these methods, as the number of dimensions rises, the latent manifolds become progressively more curved. Thus, increasing the number of dimensions intensifies the geometric distortions caused by transforming the flat initial manifolds into curved ones. This observation can be explained by the increase in divergent information when $D$ increases, which translates into more curved manifolds for encoding more divergent information. Such geometric distortions imply that the final node representations cannot be easily exploited by downstream tasks.
 
\textit{(2) The difference between the LID and ID is high even when $D$ is small:} This behavior is observed in GATNE, X-GOAL, and HMGE, which learn latent codes that reside in highly curved manifolds interdependently of $D$. X-GOAL and GATNE employ graph transformation techniques, which can corrupt the topological structures and cause strong divergence among the initial views even when $D$ is small. HMGE leverages hierarchical aggregations and thus can capture convoluted relations across dimensions. We argue that hierarchical graph dimension embedding gives birth to curved manifolds. Similar to image datasets, encoding pixel representations hierarchically has a characteristic curvature increase \cite{kaufman2022curved,ansuini2019intrinsic}.

While first-category methods can prevent geometric distortions for low-dimensional graphs, both categories still exhibit such distortions when embedding \textit{high-dimensional} graphs. \methodname is our response to this challenge; it is a multiplex graph embedding approach that yields node representations constrained to flat and low-dimensional manifolds.

\section{The Proposed \methodname Approach}

In this section, we address the problem of geometric distortions that negatively affect the latent manifolds of multiplex graph embedding methods. Specifically, we answer the question: \textit{How to encode the hierarchical latent dimensions of a multiplex graph without causing geometric distortions?} Our goal is to capture complex, yet relevant relations across the graph dimensions in a way that prevents high geometric distortions, especially in high-dimensional graphs where such issues are more prominent. To this end, we introduce \methodname, a novel approach to multiplex graph embedding that hierarchically encodes the graph dimensions while projecting node representations into hyperbolic spaces. In this context, we argue that our method not only effectively encodes hierarchical structures within-dimension but also hierarchical between-dimension relations. An example of hierarchical relations in multiplex graphs can be found in Appendix \ref{app:hierarchical_relations}.

We first introduce some notations used throughout the paper. We consider a $D$-dimensional multiplex graph $G$, defined as a set of $D$ graphs $G = (G_1, \: G_2, \: \dots, \: G_D)$, where $G_d = (V, \: A_d, \: X)$ is a graph with $N$ nodes from the set $V = \left \{ v_1, v_2, \dots, v_N \right \}$, $A_d \in \{0, 1\}^{N \times N}$ is the adjacency matrix and $X \in \mathbb{R}^{N \times F}$ is the node features matrix. Each graph $G_d$ represents a dimension of the multiplex graph $G$. Dimensions share the same nodes and features, but differ in their adjacency matrices. Given $G$, the goal is to learn an $M$-dimensional vector representation $z_i \in \mathbb{R}^M$ for each node $v_i \in V$, forming a matrix of node embeddings $Z \in \mathbb{R}^{N \times M}$.

\subsection{Hyperbolic Multiplex Graph Embedding}

Euclidean-based Graph Convolutional Network (GCN) conducts a series of $L$ message-passing operations according to:
\begin{equation}\label{eq:gcn}
    H^{(l)}_d = \sigma\big(\hat{A}_d \, H^{(l-1)} \, W^{(l)}_d\big),
\end{equation}
where $l$ is the layer index, $\hat{A}_d = \hat{\Delta}^{-\frac{1}{2}}_d (A_d + I) \hat{\Delta}^{-\frac{1}{2}}_d$, $\hat{\Delta}_d = \text{diag}(\hat{A}_d^{(l-1)} \, \boldsymbol{1}_{N})$, $\boldsymbol{1}_{N} \in \mathbb{R}^{N}$ is a vector of ones, $W^{(l)}_d$ is the weight matrix of the GCN, and $\sigma$ is an activation function. The propagation rule in hyperbolic spaces applies mapping functions to project from Euclidean space to Riemannian manifolds and vice versa \cite{liu2019hyperbolic} as given by:
\begin{equation}\label{eq:hyperbolic_gcn}
    H^{(l)}_d = \sigma\big( \exp_{\textbf{x}}\big( \hat{A}_d \log_{\textbf{x}}( H^{(l-1)})  W^{(l)}_d \big) \big),
\end{equation}
where $\exp_{\textbf{x}}$ and $\log_{\textbf{x}}$ are the mapping functions, and $\textbf{x}$ is a chosen point in the Riemannian manifold. The logarithmic map $\log_{\textbf{x}}$ projects the points from the Riemannian manifold to the Euclidean space, which allows to compute linear operations (matrix multiplications in this case). After that, the points are mapped back to the Riemannian manifold with the exponential map $\exp_{\textbf{x}}$.

Prior works have demonstrated that in a hyperbolic space of constant negative curvature, the embedding of a hierarchical structure can be achieved with lower dimensional distortion compared to Euclidean space \cite{sarkar2011low, nickel2017poincare}. Specifically, a hyperbolic space can represent hierarchical relationships more efficiently due to its exponential growth in volume with distance from the origin, aligning with the branching factor of hierarchical structures. In an embedding procedure, the topology of a hyperbolic space is defined by the mapping functions $\log_{\textbf{x}}$ and $\exp_{\textbf{x}}$. In this paper, we consider the Poincaré Ball and Lorentz definitions:

\paragraph{\textbf{(a) Poincaré Ball Model: }} This model defines a Riemannian manifold with the following mapping functions:
\begin{equation}
    \exp_{\textbf{x}}(h) = \textbf{x} \oplus \Big(\tanh\big(\frac{\lambda_{\textbf{x}} \norm{h}}{2}\big) \, \frac{h}{\norm{h}}\Big),
\end{equation}
\begin{equation}
    \log_{\textbf{x}}(h) = \frac{2}{\lambda_{\textbf{x}}} \, \text{arctanh} \left( \norm{ -\textbf{x} \oplus h } \right) \, \frac{-\textbf{x} \oplus h}{\norm{ -\textbf{x} \oplus h }},
\end{equation}
where $\lambda_{\textbf{x}} = \frac{2}{1 - ||x||^2}$ and the symbol $\oplus$ represents the Möbius addition operation \cite{ganea2018hyperbolic}. Following previous work \cite{ganea2018hyperbolic}, we set \textbf{x} to the origin point.

\paragraph{\textbf{(b) Lorentz Model: }} It is a model that exhibits better empirical performance than the Poincaré Ball Model \cite{liu2019hyperbolic}. Unlike previous models, it operates on $(M + 1)$-dimensional vectors. We first define the Minkowski inner product: $\langle  \textbf{x} \; , \; h \rangle_{\mathcal{L}} = -x_0 y_0 + \sum_{i = 1}^M x_n h_n$. Then, the mapping functions for the Lorentz model are:
\begin{equation}
    \exp_{\textbf{x}}(h) = \text{cosh}(\norm{h}_{\mathcal{L}}) \, \textbf{x}  \, +  \, \text{sinh}(\norm{h}_{\mathcal{L}})  \, \frac{h}{\norm{h}_{\mathcal{L}}},
\end{equation}
\begin{equation}
    \log_{\textbf{x}}(h) = \frac{\text{arcosh} \left(- \langle  \textbf{x} , h \rangle_{\mathcal{L}} \right)}{\sqrt{\langle  \textbf{x} , h \rangle_{\mathcal{L}}^2 - 1}} \left( h + \langle  \textbf{x} , h \rangle_{\mathcal{L}} \textbf{x} \right).
\end{equation}
In this case, we use $\textbf{x} = (1, 0, \dots, 0) \in \mathbb{R}^{M + 1}$ to account for the additional coordinate.

The Poincaré and Lorentz models allow learning representations that reside in hyperbolic spaces, leading to embeddings with lower geometric distortion. However, the high dimensionality of multiplex graphs and their hierarchical nature still affect the geometry of latent manifolds. In particular, as shown in Section \ref{seq:geo_study}, the increase of divergent information and dilution of relevant information which accompanies the increase in the dimensionality of multiplex graphs lead to geometric distortions in latent spaces. To address this issue, we introduce an additional operation in each layer of \methodname that summarizes the dimensions of the multiplex graph into a smaller and more informative set of dimensions. This mechanism is called Hierarchical Aggregations and consists of a multi-step combination of the graph adjacency matrices. Let $D_{l-1}$ be the number of input dimensions of the $l$-layer (initially, $D_0 = D$). Then, the following formula combines $D_{l-1}$ adjacency matrices to form $D_{l}$ higher-order adjacency matrices:
\begin{equation}\label{eq:hierar_aggr}
    A^{(l)}_{j} = \phi\Big(\sum_{i=1}^{D_{l-1}} \alpha^{(l)}_{i, j} A^{(l-1)}_{i}\Big),
\end{equation}
where $\phi$ is an activation function, and $\alpha^{(l)}_{i, j}$ is the weight that quantifies the importance of the $i^{\text{th}}$ input dimension in the $j^{\text{th}}$ output dimension. These weights are trainable parameters of the model, and they are normalized with a softmax function before computing the hierarchical aggregation:
\begin{equation}
    \alpha^{(l)}_{i, j} = \frac{\exp{(\hat{\alpha}^{(l)}_{i, j})}}{\sum_{k = 1}^{D_{l-1}} \exp{(\hat{\alpha}^{(l)}_{k, j})}}.
\end{equation}

The next layer uses the newly computed adjacency matrices $A^{(l)}_{j}$ to refine the node embeddings. Moreover, matrices $A^{(l)}_{j}$ are once again combined into an even smaller set of dimensions, forming a multi-level hierarchy of adjacency matrices that capture latent graph structures. Finally, inside each layer $l$, node embeddings $H^{(l)}_d$ extracted from the combined dimensions are aggregated to consensus embeddings: 
\begin{equation}
    H^{(l)} = \sum_{d=1}^{D_{l-1}} \beta^{(l)}_{d} H^{(l)}_d, 
\end{equation}
where $\beta^{(l)}_{d}$ are attention weights. The final layer outputs node embeddings $Z = H^{(L)}$. With the extraction of higher-order adjacency matrices in Equation (\ref{eq:hierar_aggr}), the embedding process encodes hierarchical relations within and between the graph dimensions. In addition, stacking multiple aggregation layers allows to capture increasingly complex relevant patterns hidden in non-linear combinations of the graph dimensions. In Section \ref{sec:hypermge_geometric_study}, we conduct a geometric study that shows that hyperbolic and hierarchical embeddings alleviate the effects of geometric distortions. Before that, we give intuitions and insights on how the proposed approach can extract node representations that reside in flat low-dimensional manifolds, thus avoiding geometric distortions.

In multiplex graphs, the source of geometric distortions is two-fold. First, as a result of the increase of divergent information and dilution of relevant information, high-dimensional multiplex graphs contain complex latent structures hidden across the graph dimensions. Second, encoding a graph dimension with a complex structure leads to coarse geometric distortions in the latent space \cite{mrabahescaping,xiong2023geometric}.  We argue that \methodname addresses these two points. In \cite{sarkar2011low}, the author discusses methods for embedding tree-like structures in hyperbolic spaces, providing theoretical evidence that hyperbolic spaces can achieve embeddings with lower distortion compared to Euclidean spaces. Furthermore, it has been demonstrated that hyperbolic embeddings can significantly outperform Euclidean embeddings in terms of learning hierarchical representation with lower dimensional distortion \cite{nickel2017poincare}. Finally, the authors of \cite{ganea2018hyperbolic} extend the application of hyperbolic geometry to neural networks, providing further evidence that hyperbolic spaces can represent data with inherent hierarchical structures more efficiently than Euclidean spaces. Thus, hyperbolic embedding can effectively encode complex hierarchical topologies within graph dimensions and hierarchical relations between graph dimensions into Riemannian manifolds. On the other hand, hierarchical aggregations gradually reduce the number of graph dimensions, generating high-order adjacency matrices that capture informative and relevant latent structures while encoding complex relations between nodes \cite{hmge2023}. The synergy of hyperbolic geometry and hierarchical aggregations allows \methodname to avoid coarse geometric distortions, leading to embeddings that reside in flat, low-dimensional manifolds.

\subsection{Training Algorithm}

To train \methodname, we optimize the Deep Graph Infomax loss function \cite{velickovic2019deep}. First, we compute a graph-level representation by aggregating the node embeddings: $s = \frac{1}{N} \sum_{i = 1}^{N} z_i$. Then, we maximize the mutual information between $s$ and node embeddings from the set $Z = \{z_1, z_2, \dots, z_N\}$. To achieve this, we sample positive pairs $(s, z_i)$ from $G$, and negative pairs $(s, \hat{z}_i)$ from $\hat{G}$, a corrupted version of $G$ obtained by randomly shuffling the features. After that, we train a discriminator $\mathscr{D}$ to distinguish between positive and negative pairs. We use bilinear scoring for the discriminator: $\mathscr{D}(h_i, s) = \text{Sigmoid}(h_i \, Q \, s)$, where $Q \in \mathbb{R}^{M \times M}$ is a parameter matrix. The loss function is:
\begin{equation}\label{eq:loss_function}
    \mathcal{L} = \sum_{i=1}^{N} \log \mathscr{D}(s, z_i) + \sum_{j=1}^{N} \log ( 1 - \mathscr{D}(s, \hat{z}_j)).
\end{equation}

\begin{algorithm}[t]
\caption{\textit{The $l$-th layer of \methodname}}
\label{algo:algo}
\begin{algorithmic}[1]

\REQUIRE multiplex graph $G^{(l-1)}$, node embeddings  $H^{(l-1)}$, mapping functions $\exp_{\textbf{x}}$ and $\log_{\textbf{x}}$.
\ENSURE  multiplex graph $G^{(l)}$, node embeddings $H^{(l)}$. \newline

\FOR{$d \gets 1$ to $D_{l-1}$}
    \STATE $H^{(l)}_d \gets \sigma\left[ \exp_{\textbf{x}}\left( \hat{A}_d \log_{\textbf{x}}\left( H^{(l-1)} \right) W^{(l)}_d \right) \right]$
\ENDFOR
\STATE $H^{(l)} = \sum_{d=1}^{D_{l-1}} \beta^{(l)}_{d} \, H^{(l)}_d$ \newline

\FOR{$j \gets 1$ to $D_{l}$}
    \STATE $A^{(l)}_{j} \gets \phi(\sum_{i}^{D_{l-1}} \alpha^{(l)}_{i, j} A^{(l-1)}_{i})$
    \STATE $G^{(l)}_j \gets (V, \, A^{(l)}_{j}, \, H^{(l)})$
\ENDFOR
\STATE $G^{(l)} \gets (G_1^{(l)}, \, G_2^{(l)}, \, \dots, \, G_{D_l}^{(l)})$\newline

\STATE \textbf{return} $G^{(l)}, \, H^{(l)}$. \newline
\end{algorithmic}
\end{algorithm}

Algorithm \ref{algo:algo} summarizes the embedding layers of \methodname. The time complexity of \methodname is $\mathcal{O}\left(T L D \left(\mathcal{E} \left( M + D \right) + N M^2 \right) \right)$, where $T$ is the number of iterations, $N$ is the number of nodes, $D$ the number of input dimensions, $L$ the number of embedding layers, $M$ the size of node embeddings, and $\mathcal{E}$ is the maximum number of edges in the graph dimensions.  The memory complexity is $\mathcal{O}\left(L D M \left( N + M \right) + L D \mathcal{E} + 2 L D^{3} \right)$. Both are linear with respect to the number of nodes $N$ and the maximum number of edges $\mathcal{E}$.

\subsection{Geometric Study of \methodname}\label{sec:hypermge_geometric_study}

\begin{figure}
    \centering
    \includegraphics[width=0.32\textwidth]{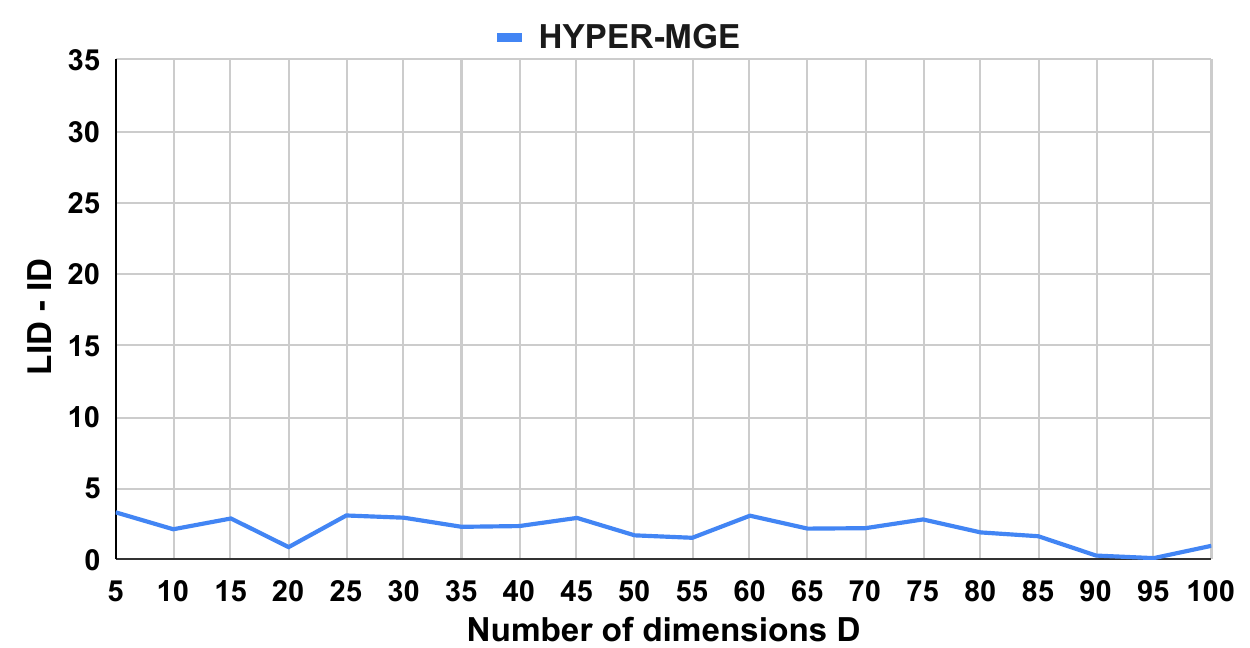}
    \caption{Results of the geometric study with \methodname.}
    \label{fig:mymodel_geo}
\end{figure}

In Figure \ref{fig:mymodel_geo}, we report the results of \methodname according to the experimental protocol of Section \ref{seq:geo_study}. We observe that the difference between the LID and ID is small (between $0$ and $4$) and does not increase when $D$ increases. This contrasts with the baselines in Figure \ref{fig:geometrical_study}, where the LID and ID tend to diverge from each other. Unlike concurrent methods, \methodname can prevent geometric distortions independently from the number of graph dimensions and learns node representations that lie in flat low-dimensional spaces. Accordingly, our approach is less affected than the baselines by the high dimensionality of the data in Figure \ref{fig:geometrical_study}. In other words, the synergy of hierarchical and hyperbolic embeddings permits \methodname to counter the increase in divergent information and dilution of relevant information when the number of dimensions increases. In the next section, we show that our contributions translate to significant empirical improvements compared to the state-of-the-art.

\section{Experiments}\label{sec:experiments}

We conduct an empirical evaluation to demonstrate the suitability of the proposed approach for high-dimensional multiplex graph embedding. We compare \methodname \footnote{The code can be found on GitHub: \url{https://github.com/abdouskamel/HYPER-MGE}.} against the state-of-the-art methods discussed in Section \ref{sec:geometric_study_protocol}, namely: MultiVERSE, GATNE, SSDCM, DMGI, HDMI, MGDCR, DMG, X-GOAL, and HMGE.

\subsection{Datasets}

\begin{table}
\centering
\resizebox{0.8\linewidth}{!}{\begin{tabular}{ c|c|c|c|c } 
\hline
Dataset & \#Dimensions& \#Nodes & \#Edges & \# Classes \\ 
\hline
BIOGRID & 28 & 4,503 & 311,645 & 4 \\
\hline
DBLP-Authors & 10 & 5,124 & 33,250 & 4 \\
\hline
IMDB & 8 & 3,000 & 224,984 & 3 \\
\hline
STRING-DB & 7 & 4,083 & 4,923,554 & 3 \\ 
\hline
\end{tabular}}
\caption{Real-world data statistics.}
\label{table:data_statistics}
\end{table}

In this section, we describe the real-world datasets used in the experiments. Table \ref{table:data_statistics} summarizes their statistics\footnote{We draw the attention of the reader to the fact that there is a scarcity of labeled high-dimensional multiplex graphs. In this setting, the datasets used in our experiments are fairly sizeable to conduct an objective evaluation.}.

\textit{BIOGRID} and \textit{STRING-DB} are protein-protein interaction graphs collected from \cite{oughtred2021biogrid} and \cite{szklarczyk2019string}, respectively. The nodes are proteins and the edges are interactions between the proteins. The edges in each dimension are inferred by different experimental protocols (e.g., the biochemical effect of one protein on another). In BIOGRID, node classification labels indicate the species from which the protein is extracted. In STRING-DB, labels represent protein families. 

\textit{DBLP-Authors} is an academic graph assembled from AMiner \cite{tang2016aminer}. Authors of research papers are depicted as nodes, and an edge indicates a co-authorship relation between two authors. Dimensions represent various conferences and journals where authors have co-written papers. Classification labels indicate the authors' research areas (thus, in this dataset, nodes can be part of multiple classes). 

\textit{IMDB}\footnote{\url{https://www.imdb.com/interfaces/}} is a multiplex graph where nodes are movies, and edges indicate that at least a person has participated in both movies. Different dimensions represent different roles: actors, directors, producers, etc. Nodes are labeled with the movie genre.

\subsection{Evaluation Protocol \& Parameter Settings}

We evaluate our approach on two downstream tasks: link prediction and node classification. To perform link prediction on the hyperbolic embeddings of \methodname, we use the Fermi-Dirac decoder \cite{nickel2017poincare} to compute probability scores between edges. It is a generalization of the sigmoid function to hyperbolic spaces as described by:
\begin{equation}
    \mathcal{P}(z_i, z_j) = \left[ e^{{\left(\text{artanh}\left(\norm{- z_i \oplus z_j}\right)^2-r\right)/t}} + 1 \right]^{-1}
\end{equation}
where $r$ and $t$ are hyper-parameters. For other baselines, we use the standard dot product $\text{Sigmoid}(ZZ^T)$, since their representations are Euclidean. We report the area under the ROC curve (AUC-ROC) and average precision (AP). For node classification, we first map the hyperbolic embeddings to the Euclidean space and then run logistic regression \cite{chami2019hyperbolic}. We assess the results with F1-Macro and F1-Micro.

For \methodname, we set the embedding size to $96$ and use 2 hierarchical aggregation layers, with ReLU as the activation function $\phi$ and Leaky ReLU as $\sigma$. Based on previous work \cite{chami2019hyperbolic,liu2019hyperbolic}, we use the Lorentz model as the Riemannian manifold and set the Fermi-Dirac decoder hyperparameters to $r = 2$ and $t = 1$. We use Adam with a learning rate of $0.001$ and a weight decay of $10^{-5}$. We train for $1,000$ epochs with early stopping after $20$ iterations. These settings are kept constant on all datasets. Finally, we run our model 5 times and report the mean and standard deviation. For the baselines, we only report the best result among 5 trials.

\subsection{Evaluation Results}

\begin{table}[t]
\centering
\resizebox{0.94\linewidth}{!}{
\begin{tabular}{ c|c|c|c|c|c|c|c|c } 
\hline
Dataset & \multicolumn{2}{|c|}{BIOGRID} & \multicolumn{2}{|c|}{DBLP-Authors} & \multicolumn{2}{|c|}{IMDB} & \multicolumn{2}{|c}{STRING-DB} \\

\hline 
Metrics & AUC & AP & AUC & AP & AUC & AP & AUC & AP \\
\hline
MultiVERSE & 50.14 & 50.07 & 54.54 & 52.38 & 48.34 & 49.15 & 50.00 & 50.00 \\
GATNE & 38.63 & 42.16 & 43.57 & 44.34 & 37.54 & 41.05 & 48.15 & 47.89 \\
SSDCM & \textit{OOM} & \textit{OOM} & 62.74 & 58.54 & 53.84 & 54.48 & 50.40 & 50.20 \\
DMGI & 50.00 & 50.00 & 50.00 & 50.00 & 52.84 & 52.66 & 54.62 & 52.69 \\
HDMI & 47.10 & 47.06 & 60.64 & 59.36 & 50.95 & 50.76 & 54.46 & 53.96 \\
DMG & \textit{OOM} & \textit{OOM} & 44.36 & 51.97 & 61.91 & 57.09 & 52.06 & 54.81 \\
HMGE & \underline{73.34} & 69.07 & \underline{67.31} & \underline{67.21} & 57.42 & 55.77 & \underline{65.46} & \underline{62.84} \\
MGDCR & 47.33 & 49.25 & 66.57 & 66.64 & 61.13 & 58.08 & 64.88 & 62.28 \\
X-GOAL & 71.22 & \underline{69.17} & 66.08 & 64.13 & \underline{63.78} & \underline{63.80} & 55.84 & 52.03 \\
\hline
\methodname (mean) & \textbf{74.70} & \textbf{72.34} & \textbf{67.66}  & \textbf{71.49} & \textbf{77.11} & \textbf{77.78} & \textbf{77.29} & \textbf{76.09} \\
\methodname (std) & 0.44 & 0.32 & 0.15 & 0.10 & 0.37 & 0.39 & 0.36 & 0.39 \\
\hline
\end{tabular}}
\caption{Link prediction results. 
}
\label{table:link_prediction}
\end{table}

\begin{table}[t]
\centering
\resizebox{\linewidth}{!}{
\begin{tabular}{ c|c|c|c|c|c|c|c|c } 
\hline
Dataset & \multicolumn{2}{|c|}{BIOGRID} & \multicolumn{2}{|c|}{DBLP-Authors} & \multicolumn{2}{|c|}{IMDB} & \multicolumn{2}{|c}{STRING-DB} \\

\hline 
Metrics & F1-Ma & F1-Mi & F1-Ma & F1-Mi & F1-Ma & F1-Mi & F1-Ma & F1-Mi \\
\hline
MultiVERSE & 95.76 & 95.84 & 57.9 & 60.72 & 41.3 & 41.3 & 46.68 & 47.75 \\
GATNE & 95.57 & 95.66 & 58.12 & 71.34 & 42.52 & 42.31 & 70.1 & 72.11 \\
SSDCM & \textit{OOM} & \textit{OOM} & 57.67 & 71.70 & 24.78 & 33.59 & 61.13 & 65.84 \\
DMGI & 32.02 & 32.82 & 54.49 & 62.36 & 38.2 & 38.2 & 65.61 & 67.62  \\
HDMI & 38.23 & 40.03 & 57.1 & 70.8 & 38.9 & 39.5 & 72.03 & 73.94 \\
DMG & \textit{OOM} & \textit{OOM} & 59.12 & 71.17 & 23.97 & 33.66 & 41.28 & 52.69 \\
HMGE & \underline{98.17} & \underline{98.24} & 57.52 & \underline{71.76} & \underline{43.02} & \underline{43.16} & \underline{80.33} & \underline{82.08} \\
MGDCR & 28.65 & 33.72 & 60.04 & 66.84 & 30.74 & 33.67 & 20.50 & 41.61 \\
X-GOAL & 73.36 & 73.37 & \underline{63.72} & 70.83 & 34.63 & 35.31 & 76.34 & 76.27 \\
\hline
\methodname (mean) & \textbf{98.46} & \textbf{98.50} & \textbf{64.82} & \textbf{72.15}  & \textbf{47.43} & \textbf{47.48} & \textbf{84.42} & \textbf{85.26} \\
\methodname (std) & 0.16 & 0.16 & 0.22 & 0.25 & 0.22 & 0.16 & 0.57 & 0.50 \\
\hline
\end{tabular}}
\caption{Node classification results.
}
\label{table:node_classification}

\vspace{-10pt}
\end{table}


\textbf{Link Prediction.} Table \ref{table:link_prediction} shows a comparison between \methodname and the baselines on the task of link prediction. \textit{OOM} indicates that the method has run out of memory during training. In all tables, the best result is highlighted in bold and the second best is underlined. As we can see from  Table \ref{table:link_prediction}, \methodname consistently outperforms all methods by a significant margin. In addition, We observe that the standard deviation is small, which indicates that the results of HYPER-MGE are consistent. The two most competitive baselines are X-GOAL and HMGE. X-GOAL is not suitable for high-dimensional multiplex graphs, because its embedding module is based on a single and linear aggregation step. On the other hand, HMGE leverages hierarchical aggregations but is not equipped with a mechanism to counter the effects of geometric distortions. The improvements brought by \methodname illustrate the benefits and the importance of our contributions. In addition, the results suggest that the geometric study on synthetic data in Section \ref{seq:geo_study} translates to practical improvements on real-world datasets.


\textbf{Node Classification}. Table \ref{table:node_classification} shows the results of node classification, where \methodname outperforms the baselines in all datasets. Furthermore, computing the p-values of the paired t-test of HYPER-MGE and the most competitive approach (HMGE) in a population of $5$ samples for each model confirms the significance of our results. We remark that random walk-based approaches (MultiVERSE and GATNE) underperform compared to \methodname. This is because these methods employ a sub-optimal local optimization algorithm to encode long-range dependencies spanning multiple dimensions. Furthermore, we can see that GNN methods (SSDCM, DMGI, HDMI, DMG, MGDCR, and X-GOAL) are less competitive than HMGE and \methodname, which harness hierarchical aggregations on top of GNNs. This shows that hierarchical embeddings lead to empirical improvements in high-dimensional datasets. Finally, hyperbolic embeddings allow \methodname to further increase the prediction accuracy compared to HMGE. In particular, hyperbolic embeddings alleviate geometric distortions in latent spaces. As illustrated in Section \ref{sec:hypermge_geometric_study}, this results in node representations that lie in flat and low-dimensional spaces suitable for downstream tasks like node classification.

\subsection{Ablation Study}\label{sec:ablation}

To demonstrate the synergy between hierarchical and hyperbolic embeddings, we conduct an ablation study with \methodname and a modified version of X-GOAL that incorporates the same hyperbolic multiplex embedding mechanism. More precisely, we have changed the code of X-GOAL to equip it with hyperbolic modules using different Riemannian manifolds to encode the graph. Lorentz and Poincaré are hyperbolic spaces, while Euclidean is equivalent to a standard GNN. Here, we draw the reader's attention to the fact that incorporating hyperbolic modules into existing approaches is not a straightforward task, as it requires a careful reimplementation of the existing code. Furthermore, not all existing methods can be easily modified to incorporate hyperbolic modules. In these settings, we choose to compare with X-GOAL because \textbf{(i)} there is a possibility to reimplement X-GOAL in order to incorporate hyperbolic modules and \textbf{(ii)} it is among the most competitive baselines (see Table \ref{table:link_prediction}), so we claim that the comparison with hyperbolic X-GOAL extends the results of our study to other competing algorithms. 

\begin{table}
\centering
\resizebox{\linewidth}{!}{\begin{tabular}{ c|c|c|c|c|c|c|c|c } 
\hline
Dataset & \multicolumn{2}{|c|}{BIOGRID} & \multicolumn{2}{|c|}{DBLP-Authors} & \multicolumn{2}{|c|}{IMDB} & \multicolumn{2}{|c}{STRING-DB} \\

\hline 
Metrics & AUC & AP & AUC & AP & AUC & AP & AUC & AP \\
\hline
X-GOAL-Euclidean & \underline{71.22} & \underline{69.17} & 66.08 & 64.13 & 63.78 & 63.80 & 55.84 & 52.03 \\

\methodname-Euclidean & 53.12 & 54.89 & 65.76 & 66.31 & 58.93 & 56.58 & 70.93 & 71.15 \\
\hline
X-GOAL-Poincaré & 58.04 & 56.65 & 59.02 & 61.17 & 62.58 & 62.29 & 53.73 & 54.89 \\

\methodname-Poincaré & 53.62 & 54.03 & \textbf{75.12} & \textbf{77.37} & \underline{69.20} & \underline{69.57} & \underline{70.18} & \underline{69.27} \\
\hline
X-GOAL-Lorentz & 57.10 & 54.55 & 56.73 & 55.39 & 60.79 & 60.33 & 56.69 & 55.28 \\

\methodname-Lorentz & \textbf{74.70} & \textbf{72.34} & \underline{67.66} & \underline{71.49} & \textbf{77.11} & \textbf{77.78} & \textbf{77.29} & \textbf{76.09} \\
\hline
\end{tabular}}
\caption{Ablation of the hyperbolic module. 
}
\label{table:ablation_hyperbolic}
\vspace{-15pt}
\end{table}

Table \ref{table:ablation_hyperbolic} shows the ablation results of the hyperbolic module. In most datasets, \methodname-Lorentz yields the best scores, although in DBLP-Authors it is outperformed by \methodname-Poincaré. Previous work \cite{liu2019hyperbolic} suggests that the Lorentz model generally achieves better performance on downstream tasks, which is consistent with the results of this ablation study. On the other hand, X-GOAL does not benefit from hyperbolic spaces, since X-GOAL-Euclidean performs significantly better than the Poincaré and Lorentz versions. These results showcase the inadequacy of straightforward hyperbolic encoding for multiplex graph embedding. Indeed, for X-GOAL, hyperbolic encoding does not bring any improvement; instead, it deteriorates the performance. We conclude that the hyperbolic embedding module is not sufficient to learn reliable node representations. It is the synergy between hierarchical aggregations and hyperbolic embedding that allows to learn complex relations between the graph dimensions while constraining the latent codes to flat low-dimensional manifolds. For this reason, \methodname harnesses both mechanisms to tackle the challenges of high-dimensional multiplex graphs.

To further illustrate the benefits of both hierarchical and hyperbolic embeddings, Table \ref{table:ablation_hierarchical_aggregation} illustrates an ablation study on the hierarchical aggregation layers and the combination weights $\alpha^{(l)}_{i,j}$. We compare the proposed approach with two altered models: \textbf{(i)} The first model (\textit{Weights Ablation}) does not contain the weights $\alpha^{(l)}_{i,j}$ that quantify the importance of each dimension. \textbf{(ii)} The second model (\textit{Layers Ablation}) does not employ hidden hierarchical layers to generate latent multiplex graphs. The results show that \methodname significantly improves the prediction accuracy of both altered models. In fact, the weights $\alpha^{(l)}_{i,j}$ are necessary to adjust the contribution of each dimension to the generated latent multiplex graphs. Besides, the hierarchical layers allow \methodname to gradually reduce the number of dimensions of the multiplex graphs while extracting high-level latent structures that improve the quality of node representations.

\begin{table}
\centering
\resizebox{\linewidth}{!}{\begin{tabular}{ c|c|c|c|c|c|c|c|c } 
\hline
Dataset & \multicolumn{2}{|c|}{BIOGRID} & \multicolumn{2}{|c|}{DBLP-Authors} & \multicolumn{2}{|c|}{IMDB} & \multicolumn{2}{|c}{STRING-DB} \\ 
\hline
Metrics & AUC & AP & AUC & AP & AUC & AP & AUC & AP \\
\hline
Weights Ablation & \underline{70.91} & \underline{68.57} & 59.16 & 61.9 & 67.28 & 67.75 & 74.21 & \underline{74.02} \\
Layers Ablation & 65.34 & 63.62 & \underline{60.05} & \underline{62.32} & \underline{71.85} & \underline{72.01} & \underline{75.03} & 73.77 \\
\hline
\methodname & \textbf{74.70} & \textbf{72.34} & \textbf{67.66} & \textbf{71.49} & \textbf{77.11} & \textbf{77.78} & \textbf{77.29} & \textbf{76.09} \\
\hline
\end{tabular}}
\caption{Ablation of the hierarchical aggregation module. 
}
\label{table:ablation_hierarchical_aggregation}
\end{table}

\subsection{Intrinsic Dimensionality During Training}

\begin{figure}
\centering
\begin{subfigure}{0.45\linewidth}
    \centering
    \includegraphics[width=\textwidth]{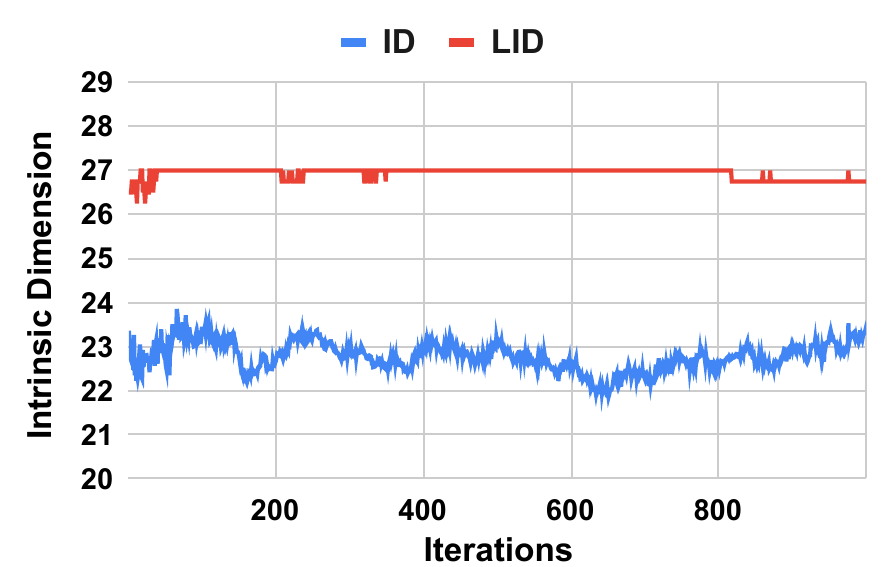}
    \caption{X-GOAL}\label{fig:xgoal_idlid}
\end{subfigure}
\begin{subfigure}{0.45\linewidth}
    \centering
    \includegraphics[width=\textwidth]{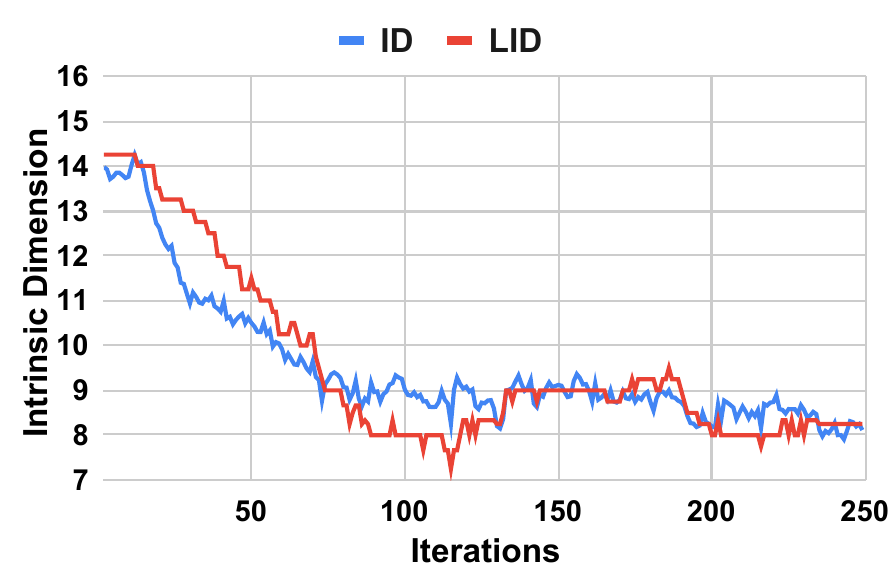}
    \caption{\methodname}\label{fig:mymodel_idlid}
\end{subfigure}

\caption{Evolution of the ID and LID metrics of X-GOAL and \methodname on BIOGRID.}
\label{fig:real_data_idlid}
\vspace{-11pt}
\end{figure}

Figure \ref{fig:real_data_idlid} shows the variations of the intrinsic dimensions during the training of \methodname and X-GOAL on the BIOGRID dataset. We monitor the ID and LID metrics after each training iteration. The results are representative of the other datasets, therefore, to save space, we chose only to include BIOGRID. We observe that the values with \methodname are close throughout training, which implies that the representations converge to a relatively flat low-dimensional space. On the other hand, as suggested by the difference between the two metrics, X-GOAL learns more curved latent manifolds. This phenomenon is caused by the high number of dimensions of BIOGRID, and the fact that X-GOAL is not equipped with mechanisms to counter the negative effects of geometric distortions. These results extend our geometric study on synthetic data to real-world datasets. Furthermore, it geometrically demonstrates the effectiveness and relevance of our contributions, and gives empirical intuitions on the reasons \methodname outperforms the baselines.

\subsection{Sensitivity Analysis}

\begin{figure}
\centering
\begin{subfigure}{0.45\linewidth}
    \centering
    \includegraphics[width=\textwidth]{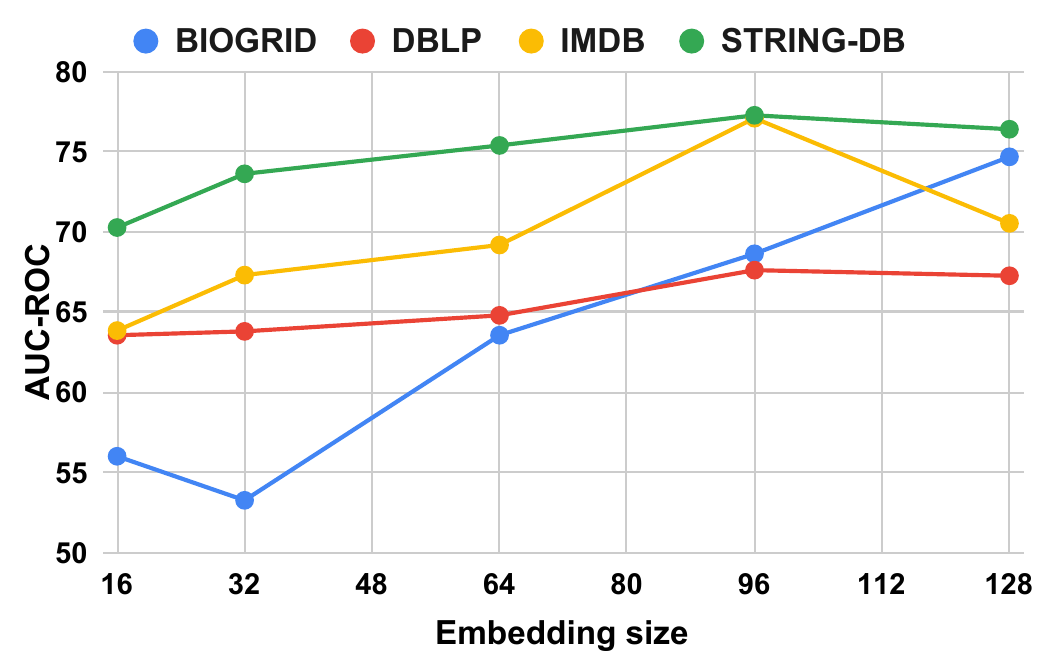}
    \caption{The size of embeddings $M$.}
    \label{fig:sensitivity_embedding_size}
\end{subfigure}
\begin{subfigure}{0.45\linewidth}
    \centering
    \includegraphics[width=\textwidth]{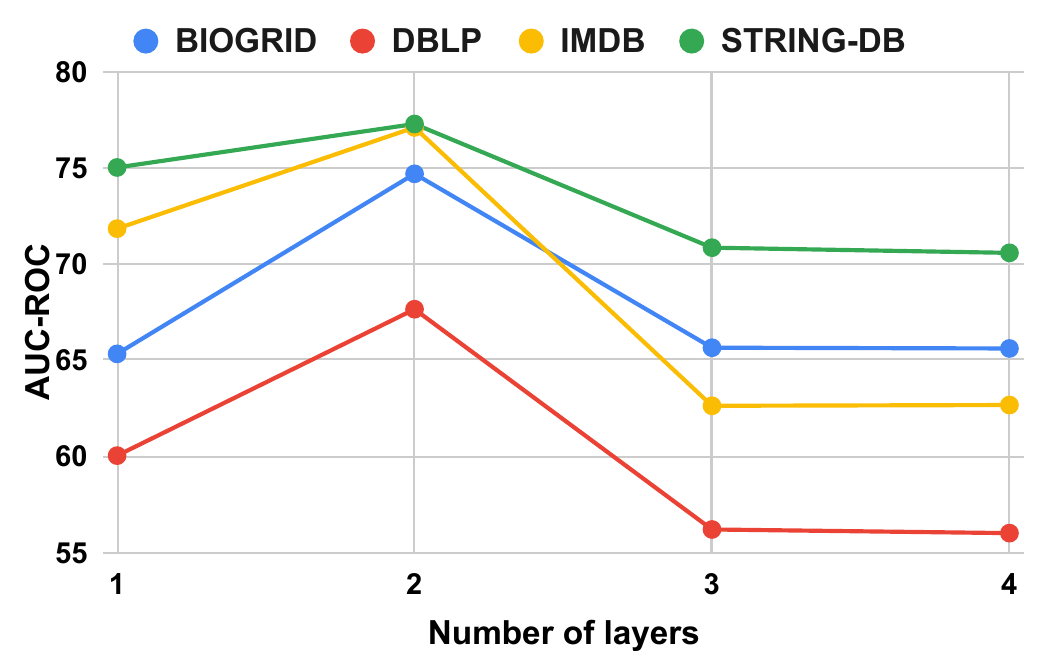}
    \caption{The number of layers $L$.}
    \label{fig:sensitivity_number_layers}
\end{subfigure}
\caption{Sensitivity analysis of \methodname.}
\label{fig:sensitivity_analysis}
\end{figure}

We study the sensitivity of \methodname to the size of node embeddings $M$ in Figure \ref{fig:sensitivity_embedding_size}. We select $M$ from the range $[16, \, 32, \, 64, \, 96, \, 128]$ and report the AUC-ROC on all datasets. On DBLP, IMBD, and STRING-DB, we see that the performance is strong for a wide range of values. In the case of BIOGRID, it requires a large embedding size to obtain satisfactory results, because it is the dataset with the highest number of dimensions. We also conduct in Figure \ref{fig:sensitivity_number_layers} a sensitivity analysis on the number of hierarchical layers $L$. We vary $L$ between 1 and 4 and evaluate the models with AUC-ROC. The graphic illustrates a similar pattern on all datasets: the performance increases when increasing $L$ from 1 to 2, and it slightly decreases when further increasing $L$. Thus, on these datasets, the optimal number of layers is 2.

\subsection{Visualizations}

\begin{figure}[t]
\centering
\begin{subfigure}{0.29\linewidth}
    \centering
    \includegraphics[width=\textwidth]{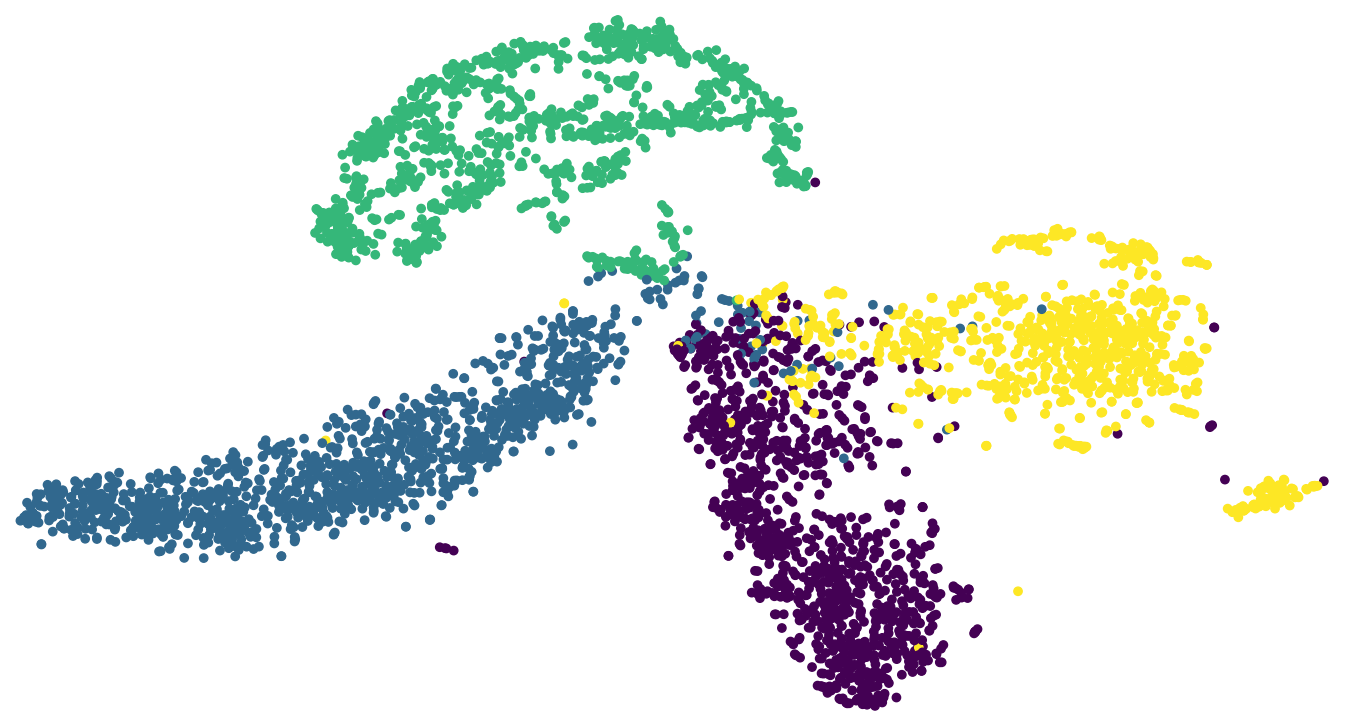}
    \caption{BIOGRID.}
\end{subfigure}
\begin{subfigure}{0.29\linewidth}
    \centering
    \includegraphics[width=\textwidth]{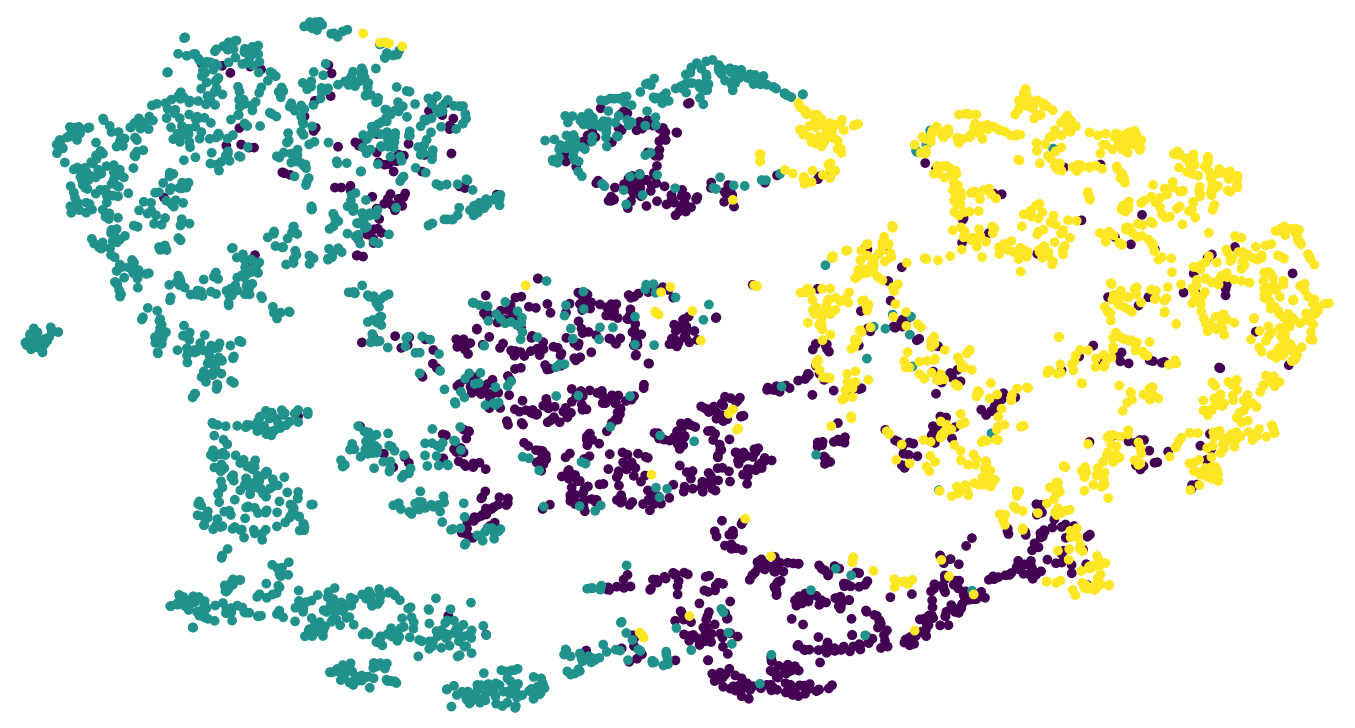}
    \caption{STRING-DB.}
\end{subfigure}
\caption{T-SNE visualization of \methodname's embeddings.}
\label{fig:embeddings_viz}
\end{figure}

Figure \ref{fig:embeddings_viz} shows t-SNE visualizations of the node embeddings extracted by \methodname on BIOGRID and STRING-DB. We observe a pronounced separation between the nodes based on the class labels. This separation can be explained by the capacity of our approach in projecting high-dimensional curved manifolds into low-dimensional flat subspaces suitable for downstream tasks. 

\section{Conclusion}

This paper investigates the problem of high-dimensional multiplex graph embedding from a geometric perspective. We find that, as the number of dimensions increases, the embedding spaces can be subjected to coarse geometric distortions that give birth to highly curved low-dimensional manifolds. This problem, which has been previously overlooked in the literature, hinders the quality of node representations for downstream tasks. To address this issue, we propose \methodname, a multiplex graph embedding method based on hierarchical and hyperbolic embedding. \methodname projects the points into Riemannian manifolds and hierarchically extracts representations that reside on flat low-dimensional spaces. Experiments on real-world high-dimensional multiplex graphs show improvements over state-of-the-art approaches. In addition, ablation studies provide evidence of the synergy between hierarchical aggregations and hyperbolic embedding.

\appendix
\section*{Appendices}

\section{Synthetic Data Generation}\label{app:synthetic_data}

The generation process is based on a modified version of the stochastic block model \cite{holland1983stochastic}. Let $G$ be a multiplex graph with $N$ nodes, $D$ dimensions, and $K$ clusters. First, we randomly assign the $N$ nodes to the $K$ clusters, so that the size of each cluster is uniformly sampled between a minimum and a maximum value. After that, based on the cluster assignments, we generate the dimensions by sampling two kinds of links:

\paragraph{\textbf{(a) Between-Cluster Links:}} These links connect nodes that are not part of the same cluster. For each dimension $d$, we perform the stochastic block model and sample between-cluster links, such that the probability of the existence of a link between two nodes is $p_{out}$. This process ensures that between-cluster regions are significantly sparser than within-cluster regions.

\paragraph{\textbf{(b) Within-Cluster Links:}} These links connect nodes that are part of the same cluster. The goal is to generate denser regions than previously so that embedding methods are capable of regrouping the nodes. However, we also want the clusters to be sparser as $D$ increases. For each cluster $k$, we first sample a spread factor $SF_{k}$ between a minimum and maximum value. The spread factor $SF_{k}$ defines the number of dimensions in which the within-cluster links will be generated (i.e., the number of dimensions in which the cluster exists). Let $N_{k}$ be the number of nodes in cluster $k$. We split the $N_{k}$ nodes into $SF_{k}$ overlapping groups, and we associate each group with a single dimension sampled uniformly. At this point, each group represents the existence of a cluster $k$ in a dimension $d$, thus simulating the property that clusters can spread into multiple dimensions. Finally, given two nodes belonging to the same group, we sample a link between them with a probability $p_{in}$. To ensure that within-cluster regions are more dense than between-cluster regions, we set $p_{in} > p_{out}$.

We set $N = 2,000$, $K = 5$, and generate multiplex graphs with a number of dimensions $D$ between $5$ and $100$. We increment $D$ by $5$ after each generation step. We sample $p_{in}$ from $[0.1, \; 0.2]$ and $p_{out}$ from $[0.01, \; 0.02]$. The spread factor $SF_{k}$ is sampled from $[1, \; 10] \in \mathbb{N}$, depending on the number of dimensions $D$.

\section{Intrinsic Dimension Metrics}\label{app:intrinsic_dim}

We use TwoNN \cite{facco2017estimating} to estimate the Intrinsic Dimension (ID) of a latent manifold. TwoNN is a widely used estimator due to its computational efficiency. Let $X = \{x_i\}_{i=1}^N$ be a set of $N$ points with an ID equal to $\delta$. For each $x_i$, we consider $r_1(i)$ and $r_2(i)$, the distances between $x_i$ and its first and second nearest neighbors, respectively. The authors of \cite{facco2017estimating} prove that the ratio $\mu_i = r_2(i)/r_1(i)$ follows a Pareto distribution with a scale parameter of $1$ and a shape parameter of $\delta$. Thus, the probability density function $f(. | \delta)$ and the cumulative distribution function $F(. | \delta)$ are:
\begin{equation}
    f(\mu_i | \delta) = \delta \, \mu_i^{-(\delta+1)} \, 1_{[1, +\infty]} (\mu_i),
\end{equation}
\begin{equation}
    F(\mu_i) = (1 - \mu_i^{-\delta}) \, 1_{[1, +\infty]} (\mu_i).
\end{equation}

Applying simple algebra, it turns out that the value of $\delta$ can be estimated by:
\begin{equation}\label{eq:id_equation}
    \delta = \frac{\log(1 - F(\mu_i))}{\log(\mu_i)}.
\end{equation}

Let $S = \left\{ \left( \log(\mu_i), -\log(1 - F(\mu_i)) \right) \right\}_{i=1}^{N}$. Equation (\ref{eq:id_equation}) indicates that $S$ is contained in a straight line passing through the origin and with a slope of $\delta$. Thus, the value of $\delta$ can be estimated by a linear regression on the set $S$.

For the Linear Intrinsic Dimension (LID), we take inspiration from \cite{ansuini2019intrinsic} and use Principal Component Analysis (PCA) to estimate the dimension of the smallest linear space that encloses the embeddings. We select the minimal number of components that explain $90\%$ of the variance in the embedding space.

\section{Hierarchical Relations in Multiplex Graphs}\label{app:hierarchical_relations}

\begin{figure}[h]
    \centering
    \includegraphics[width=0.75\linewidth]{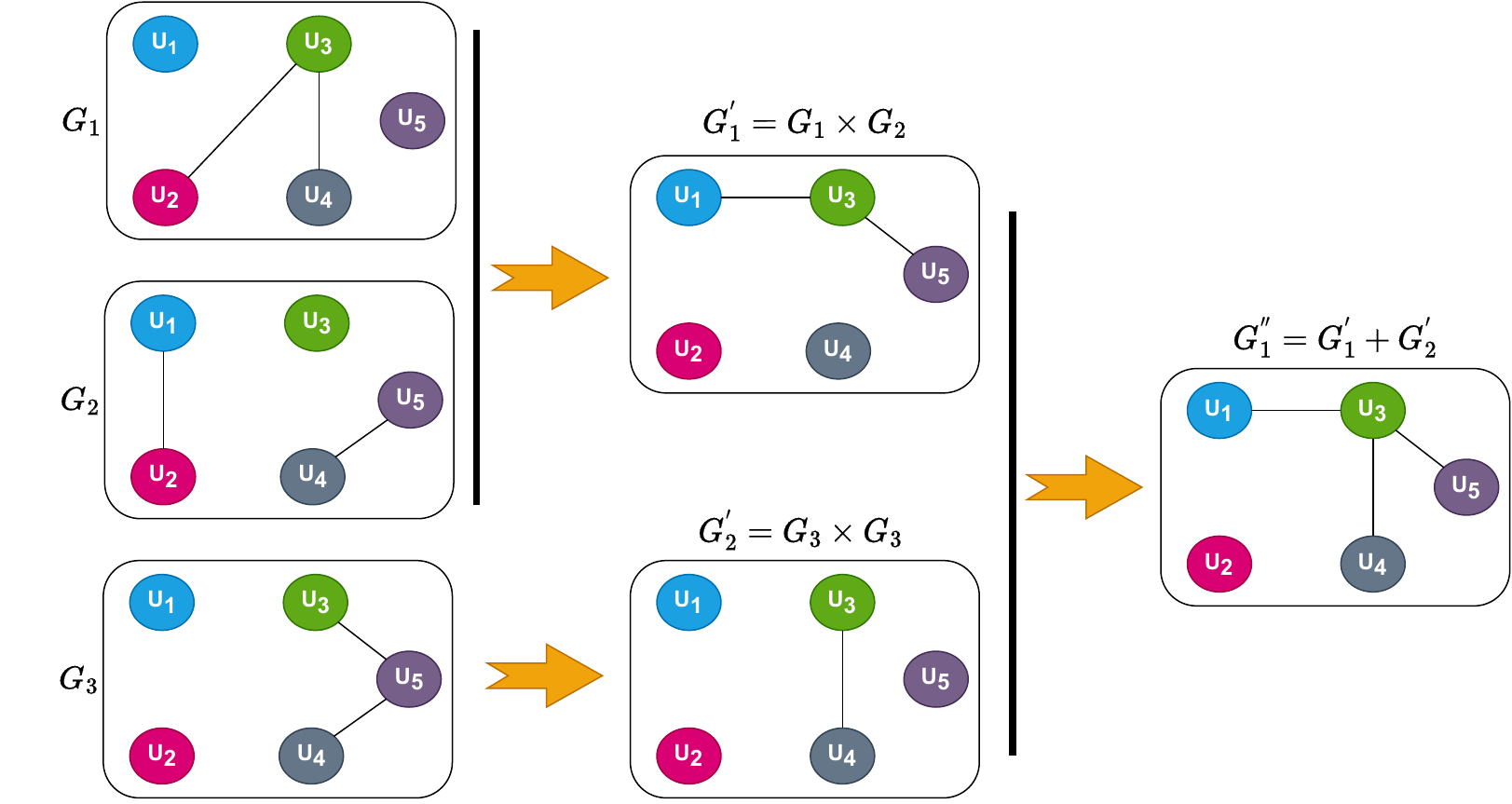}
    \caption{Example of hierarchical relations in a multiplex graph.}
    \label{fig:hierarchical_example}
\end{figure}

Hierarchical relations between the multiplex graph dimensions refer to the existence of new high-level graph dimensions that result from non-linear hierarchical combinations of the initial lower-level dimensions. Consider the example in Figure \ref{fig:hierarchical_example} where a multiplex graph has three initial dimensions $G_1$, $G_2$, and $G_3$. The initial dimensions $G_1$ and $G_2$ can be combined by the product of their adjacency matrices to generate a new dimension $G_1^{\prime}$. Nodes in $G_1^{\prime}$ are connected if they are linked by an edge from $G_1$ followed by an edge from $G_2$. Besides, another dimension $G_2^{\prime}$ can be formed by squaring the adjacency matrix of $G_3$. The new graph dimension $G_2^{\prime}$ contains meta-paths in $G_3$ of length 2. Finally, the summation of $G_1^{\prime}$ and $G_2^{\prime}$ forms a new dimension $G_1^{\prime\prime}$. We can see that these operations create a multi-level hierarchy of dimensions such that each level unravels information that was not present in the previous level. For example, the edges $u_1 \rightarrow u_3$ and $u_3 \rightarrow u_4$ do not exist in the initial dimensions. Thus, in multiplex graphs, latent information can be hidden in non-linear combinations of the dimensions. Hierarchical aggregations aim to retrieve this information by hierarchically constructing higher-semantic latent structures.

\bibliographystyle{ACM-Reference-Format}
\balance

\end{document}